\documentclass{article}

\usepackage{arxiv}

\usepackage[utf8]{inputenc} 
\usepackage[T1]{fontenc}    
\usepackage{hyperref}       
\usepackage{url}            
\usepackage{booktabs}       
\usepackage{amsfonts}       
\usepackage{nicefrac}       
\usepackage{microtype}      
\usepackage{lipsum}		
\usepackage{graphicx}
\usepackage{doi}

\title{A Contrastive Learning Scheme with Transformer innate Patches}

\author{ \href{https://orcid.org/0009-0009-2798-9991}{\includegraphics[scale=0.06]{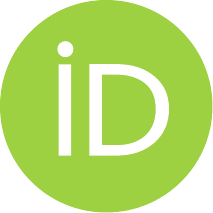}\hspace{1mm}Sander Riisøen Jyhne}\\
	Department of ICT\\
	University of Agder\\
	Grimstad, Norway \\
	\texttt{sander.jyhne@uia.no} \\
	\And
	\href{https://orcid.org/0000-0002-7742-4907}{\includegraphics[scale=0.06]{orcid.pdf}\hspace{1mm}Per-Arne Andersen} \\
	Department of ICT\\
	University of Agder\\
	Grimstad, Norway \\
	\texttt{per.andersen@uia.no} \\
        \And
	\href{https://orcid.org/0000-0001-6331-702X}{\includegraphics[scale=0.06]{orcid.pdf}\hspace{1mm}Morten Goodwin} \\
	Department of ICT\\
	University of Agder\\
	Grimstad, Norway \\
	\texttt{morten.goodwin@uia.no} \\
}

\renewcommand{\shorttitle}{\textit{arXiv} Template}

\hypersetup{
pdftitle={A template for the arxiv style},
pdfsubject={q-bio.NC, q-bio.QM},
pdfauthor={David S.~Hippocampus, Elias D.~Striatum},
pdfkeywords={remote sensing, contrastive learning, deep learning},
}

\begin{document}
\maketitle

\begin{abstract}
This paper presents Contrastive Transformer (CT), a contrastive learning scheme using the innate transformer patches. CT enables existing contrastive learning techniques, often used for image classification, to benefit dense downstream prediction tasks such as semantic segmentation. The scheme performs supervised patch-level contrastive learning, selecting the patches based on the ground truth mask, subsequently used for hard-negative and hard-positive sampling. The scheme applies to all patch-based vision-transformer architectures, is easy to implement, and introduces minimal additional memory footprint. Additionally, the scheme removes the need for huge batch sizes, as each patch is treated as an image.

We apply and test CT for the case of aerial image segmentation, known for low-resolution data, large class imbalance, and similar semantic classes. We perform extensive experiments to show the efficacy of the CT scheme on the ISPRS Potsdam aerial image segmentation dataset. Additionally, we show the generalizability of our scheme by applying it to multiple inherently different transformer architectures. Ultimately, the results show a consistent increase in mean Intersection-over-Union (IoU) across all classes.
\end{abstract}


\section{Introduction}

Segmentation of buildings from aerial images is a vital task, significantly impacting various sectors such as urban planning, disaster management, and environmental monitoring \cite{Matei2008BuildingData,Li2021JointSegmentation,Li2020InstanceKeypoints,Khoshboresh-Masouleh2020MultiscaleSensors}. The accuracy of this segmentation process is crucial, as it directly influences the estimation of population density, the development of urban planning maps, and more. However, a noticeable gap persists in existing research, especially regarding the precision of building edges. This shortfall primarily stems from challenges such as class imbalances, the high similarity between different classes in aerial image datasets, and the inherent complexity of accurately capturing the intricate details of building edges. These issues complicate tasks like fine-grained change detection and map production that rely heavily on accurate building vectorization. Consequently, manual interventions and additional costs become inevitable, underscoring the need for improved automation.

Segmenting buildings from aerial images is challenging due to the wide range of building types, including skyscrapers, residential houses, and industrial buildings. Each category has its unique shape, size, and pattern characteristics. Additional complexities come from elements like shadows, reflections, and changes in lighting that can make it harder to create accurate segmentation masks. Buildings that overlap or are partially hidden by other objects make the task even more difficult. Historically, building segmentation has been directed toward urban planning, disaster damage assessment, and change detection applications. However, the level of precision necessary for these tasks often differs from applications like building vectorization. Even though some studies have focused on the precision of building segmentation \cite{Shi2019BuildingEmbedding,Li2020InstanceKeypoints,Li2021JointSegmentation}, there is still a need for enhancing the accuracy further. This discrepancy underscores the need for continued research and development in segmentation accuracy. Moreover, the models developed must be capable of generalizing across a wide range of building types and environments, spanning urban and rural settings.

In this work, we introduce the Contrastive Transformer (CT), a supervised contrastive learning model that leverages the inherent patch-structure of the general transformer architecture \cite{Dosovitskiy2021AnScale}. Our aim with CT is to enhance the precision of existing transformer-based segmentation models through intra- and inter-image contrastive learning facilitated by the intrinsic patch structure of the transformer architecture. Simply put, intra-image contrastive learning identifies patches within a single image, while inter-image contrastive learning gathers them across multiple images. By harnessing these strategies alongside established contrastive loss functions, we aim to boost our model's performance. Depending on the image and patch size, a single image can accommodate over 20,000 patches aggregated over the transformer encoder's four stages. Patch collection is steered by ground truth masks, which enable the selection of anchor and positive samples containing a homogenous class distribution of the target class. Conversely, negative samples can display a varied class distribution, provided they exclude the target class. We implement a straightforward sampling strategy to select challenging samples for contrastive learning. Our central contribution is the Contrastive Transformer learning scheme, which, in preliminary tests, demonstrated adaptability across various vision-transformer backbones and resulted in an improved Intersection over Union (IoU) score across all classes when benchmarked against existing models. We explore these encouraging results in greater detail in the following sections of the paper. 

Section \ref{sec:related_work} provides an overview of prior work, highlighting gaps in the current literature that motivate our contributions. In Section \ref{sec:ct}, we introduce our primary contribution, the CT learning scheme. We evaluate the proposed method and compare it to previous state-of-the-art methods in aerial building segmentation in Section \ref{sec:experiments}. Finally, in Section \ref{sec:conclusion}, we summarize our findings and discuss the implications of our work. We also explore future CT directions in Section \ref{sec:future_work}.

\section{Related Work}
\label{sec:related_work}

\label{sec:related_work}

Advances in semantic segmentation often depend on developing more powerful deep neural networks. The earliest literature focused on variations of deep convolutional neural networks, such as the U-Net model \cite{Ronneberger2015U-net:Segmentation}. Despite the early success of CNN-based models, the focus has shifted to Transformer-based architectures following the release of the Vision Transformer \cite{Dosovitskiy2021AnScale}. Transformers have proven effective in semantic segmentation tasks, even though they bring along certain scalability challenges related to the self-attention mechanism \cite{DumanKeles2023OnSelf-Attention}. In the preceding years, several works have reduced the computational complexity while maintaining or increasing the accuracy. The authors of \cite{Liu2022SwinResolution} introduce a hierarchical transformer model using shifted windows, effectively reducing self-attention computation. Other works, such as \cite{Yu2022MetaFormerVision}, replace the attention-based module with a simple spatial pooling operator performing basic token mixing. Furthermore, in \cite{Li2022UniFormer:Recognition}, the authors propose a more generalizable model that uses convolutions and self-attention for vision tasks. 

While advances in architectural design play a crucial role, an equally important aspect is the optimization strategy, which can enhance the network's performance without architectural modifications. One such strategy that has shown promise is contrastive learning. Contrastive learning is a technique aiming to reduce the distance between representations of the same semantic class while increasing the distance between representations for different classes. Contrastive learning uses a similarity metric, such as cosine similarity or Euclidean distance, to evaluate the relation between representations and give feedback to the network. The latest image-based contrastive frameworks determine similarity scores using the global representation of the data, often used in image classification \cite{Chen2020ARepresentations,Dao2021Multi-LabelLearning,vandenOordDeepMind2018RepresentationCoding,He2020MomentumLearning}. Contrastive learning can also be applied using dense representations instead of global representations of the image. In this approach, the representation of a specific class derives from a group of pixels, which has shown improved performance in dense prediction tasks such as object detection and semantic segmentation \cite{Liu2021BootstrappingContrast}. However, pixel-wise dense representations' computational and memory demands present significant challenges. Our method mitigates these limitations by leveraging the existing representations in the Transformer backbone for contrastive learning, effectively enhancing performance while minimizing computational complexity and memory requirements. This strategy sets our approach apart from other methods in contrastive learning for semantic segmentation, which we will now discuss.

Various strategies have been pursued in contrastive learning for semantic segmentation, each presenting its unique strengths and limitations. One frequently employed technique involves a two-stage training process, where the initial phase focuses on pre-training the backbone with contrastive learning, and the subsequent phase fine-tunes it for segmentation. Examples of this approach can be seen in works like \cite{Zhang2021LookingLearning} and \cite{Zhao2021ContrastiveSegmentation}, which generate auxiliary labels in conjunction with ground truth labels for contrastive learning. A drawback to this approach, however, is its substantial memory consumption. In contrast, our work leverages an end-to-end training process, eliminating the need for two-stage training. Other research, including \cite{Wang2021ExploringSegmentation,Alonso2021Semi-SupervisedBank}, has also employed end-to-end training strategies, but these methods typically rely on a memory bank to store features during training, compromising efficiency. Our approach addresses this by selecting features in batches on the fly. Additionally, while the active sampling in \cite{Alonso2021Semi-SupervisedBank} uses class-specific attention modules, our method utilizes ground truth labels to choose the patches used for contrastive learning. Lastly, in \cite{Liu2021BootstrappingContrast}, a contrastive learning framework for regional learning is proposed to support semantic segmentation with end-to-end learning and active sampling. However, they sample key pixels using a class relationship graph, and hard queries are chosen based on the predicted confidence map, a methodology that differs from ours.

In aerial image segmentation, contrastive learning has also been adopted to enhance semantic representations. For instance, \cite{Huang2022ARECOGNITION} explores contrastive learning at the semantic level using the decoder output, which differs significantly from our usage of the encoder representations. Another notable example is \cite{Tang2023SemanticConsistency}, which employs a two-stage strategy incorporating a data augmentation policy. They aim to develop a semantically accurate representation in the encoder for aerial scene recognition, showcasing the adaptability and potential of contrastive learning in diverse applications within semantic segmentation.

\section{Contrastive Transformer}
\label{sec:ct}

\begin{figure}
    \centering\includegraphics[width=1.0\linewidth]{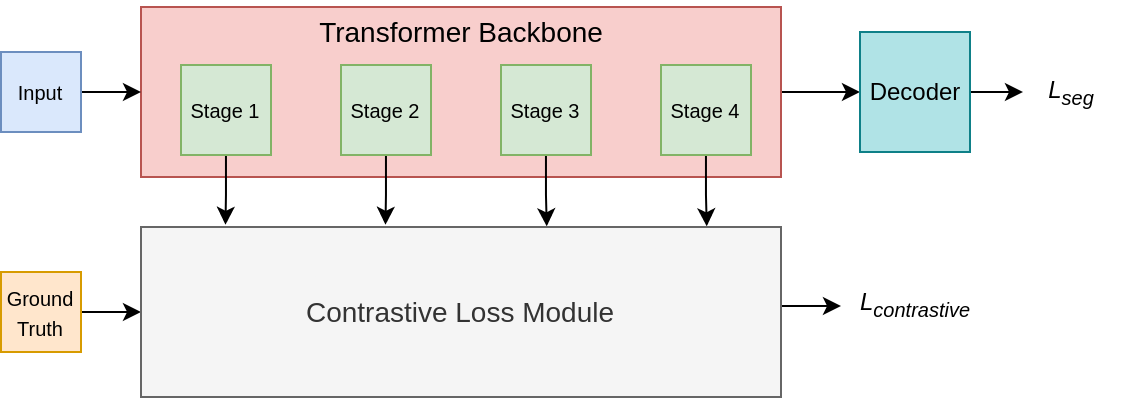}
    \caption{CT uses the innate patch representations from each encoder stage and calculates the contrastive loss between positive and negative samples using the ground truth mask.}
    \label{fig:overview}
\end{figure}

The Contrastive Transformer (CT) is an innovative patch-based contrastive learning paradigm that leverages the inherent patch-structured design of vision transformers to enhance the semantic representation of classes in an image. A high-level overview of the CT mechanism is illustrated in Figure \ref{fig:overview}. The input image is sent into the transformer backbone, where we apply contrastive learning at each encoder stage, producing contrastive feedback at multiple levels of abstraction. In each stage of the transformer, CT samples in-batch positive and negative patches, enabling the learning of a wide spectrum of robust signals and semantic representations while still being efficient. In addition to contrastive learning, a conventional segmentation loss is also applied to optimize the model for segmentation collectively. Despite its straightforward design, the CT framework demonstrates remarkable effectiveness, making it versatile and applicable across various transformer models. Its adaptability extends to its compatibility with most image-based contrastive learning methods, enhancing performance for dense prediction tasks. In the subsequent subsections, we will delve into the specifics of the CT learning scheme, the naive sampling strategy, and the contrastive loss functions utilized in our experiments.

\subsection{Architecture}

Let $(X, Y)$ represent the training dataset, where $x \in X$ represents the training images and $y \in Y$ represents the pixel-level classes in the dataset. A segmentation network $g$ is trained to learn the mapping $g_\theta : X \rightarrow Y$, where $\theta$ represents the network's parameters. The segmentation network $g$ consists of two components, the backbone $\phi$, which maps $\phi : X \rightarrow Z$, and the decoder $\omega$, which maps $\omega : Z \rightarrow Y$. To perform patch-level contrastive learning, we attach a projection head $\psi$ in parallel to the decoder network on the $\phi$ mapping, where $\psi : Z \rightarrow F$ and $F$ is an $n$-dimensional representation of $Z$. The projection head only incurs additional computational costs during training and is removed during inference.

\begin{figure}
    \centering\includegraphics[width=1.0\linewidth]{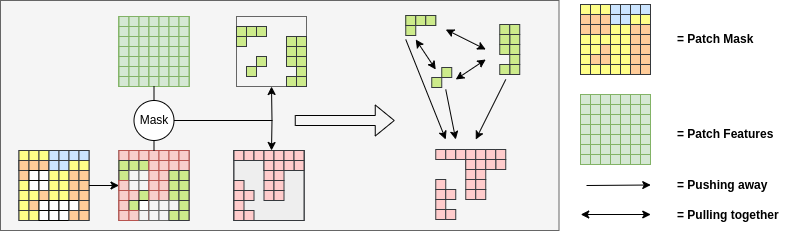}
\caption{The color-coded squares in the patch mask represent patches with a homogenous class distribution, while the white squares represent patches with a mixed class distribution including the target class. Using the ground truth mask it knows which feature representations to use as positive and negative samples. Positive feature patches consist of a uniform distribution of the target class. In contrast, negative patches may contain a mixture of classes or may be uniform as long as they exclude the target class. Patches with a mixture of classes including the target class are discarded. Ultimately, the selected positive and negative patches contribute to the contrastive loss function, pulling the representations of the positive patches closer together and pushing away the negative patch representations.}
    \label{fig:contrastive}
\end{figure}

CT collects all patch representations at each encoder stage during training and couples them with the corresponding ground truth patch. For each encoder stage $s$, we have feature patches $F_s$ and corresponding ground truth patches $G_s$ constructed with the same patch size as the feature patches. For each unique class $c$ in $G$, we sample positive patches $P_s$ and negative patches $N_s$ from $F_s$. $P_s$ are sampled from $F_s$ where $G_s$ have a homogenous class distribution of class $c$. Similarly, $N_s$ are sampled from $F_s$ where class $c$ is not in $G_s$. Figure \ref{fig:contrastive} visualizes the sampling process for positive and negative samples. The positive and negative samples are then used in a contrastive loss function.

\subsection{Sampling and Loss Functions}
Sampling strategies and loss functions form the cornerstone of contrastive learning. To start, we use a sampling strategy that sifts through and selects positive and negative samples, which are then used in the contrastive loss function. In our experiments, we adopt a simple yet effective 'naive' sampling strategy that randomizes all patches in the batch for both positive and negative samples. For negative contrastive learning, we compute the cosine similarity for a set number of positive and negative pairs, sort these pairs in descending order, and select the top 50\%. On the other hand, for positive contrastive learning, we bisect the positive samples, compute the cosine similarity, sort these in ascending order, and again select the top 50\%. Following the sampling stage, we calculate the loss via a contrastive loss function. Considering each patch as a separate image lets us use contrastive loss functions from the image classification field, broadening our options scope. Further, as each image comprises numerous patches distributed across multiple stages, there is no necessity for a large batch size to gather enough samples for contrastive learning. We implement the InfoNCE \cite{vandenOordDeepMind2018RepresentationCoding} and a custom contrastive loss function for our experiments. This custom Contrastive Loss (CL) function calculates the cosine similarity of the positive and negative samples and normalizes the results between 0 and 1, where 0 denotes dissimilarity, and 1 indicates similarity. We then employ the soft cross-entropy loss function with a smoothing parameter set to $0.1$ to compute the loss. The target is set to 0 for negative contrastive learning samples and 1 for positive contrastive learning.

\section{Experiments}
\label{sec:experiments}
This section provides evidence that CT enhances mean IoU on the ISPRS Potsdam Dataset by applying image-based contrastive loss functions on feature representations from the transformer backbone. Our experimental assessment seeks to address the subsequent key questions:

\begin{itemize}
    \item How does the performance of a straightforward approach to transformer-based contrastive learning compare to existing state-of-the-art segmentation models?
    \item Can we deduce conclusions about the model's capability to process semantic classes smaller than the smallest patch in the transformer?
    \item How does the proposed learning scheme adapt to various transformer backbones?
\end{itemize}

\subsection{Experimental Setup}

Recognizing the inherent challenges in aerial image datasets, such as large class imbalances and high similarity between different classes, we used such a dataset to test our proposed learning scheme. The capacity to distinguish between similar classes relies on the ability of the network to form robust representations of each class. This challenge is particularly important and present in our chosen dataset, namely the International Society for Photogrammetry and Remote Sensing (ISPRS) Potsdam semantic labeling dataset \cite{isprs.orgISPRSDataset} for our experiments. The dataset consists of 38 tiles, each with dimensions of 6000x6000. It classifies pixels into six distinct classes: Surface, Building, Vegetation, Tree, Car, and Clutter. Following the methodology of earlier studies \cite{Liu2020DenseClassification,Audebert2018BeyondNetworks,Liu2018SemanticNetwork}, we allocated 24 tiles for training and the remaining 14 for testing, disregarding the 'Clutter' class from the evaluation. We extracted smaller tiles of size 500x500 from the original tiles and resized them to 512x512 for training and testing. We excluded the DSMs from our experiments, intending to advance image-based research.

The performance of Contrastive Transformer (CT) was examined using three backbones, namely DCSwin (DS) \cite{wang2022aimages}, UnetFormer (UF) \cite{Wang2022UNetFormer:Imagery}, and PoolFormer (PF) \cite{Yu2022MetaFormerVision}, with the decoder originating from \cite{wang2022aimages}. All experiments used a learning rate of $8e-5$, a batch size of $32$, and deployed the AdamW optimizer. Both gradients and contrastive loss were clipped to $1.0$ to prevent contrastive learning from overtaking as the primary optimization factor. A joint loss function, constituting soft cross-entropy loss and dice loss, was used across all experiments. Every experiment underwent a training phase of 50 epochs, and the reported results represent the average of three individual runs.

\subsection{Results}
\label{sec:results}

\begin{figure}
    \centering\includegraphics[width=0.8\linewidth]{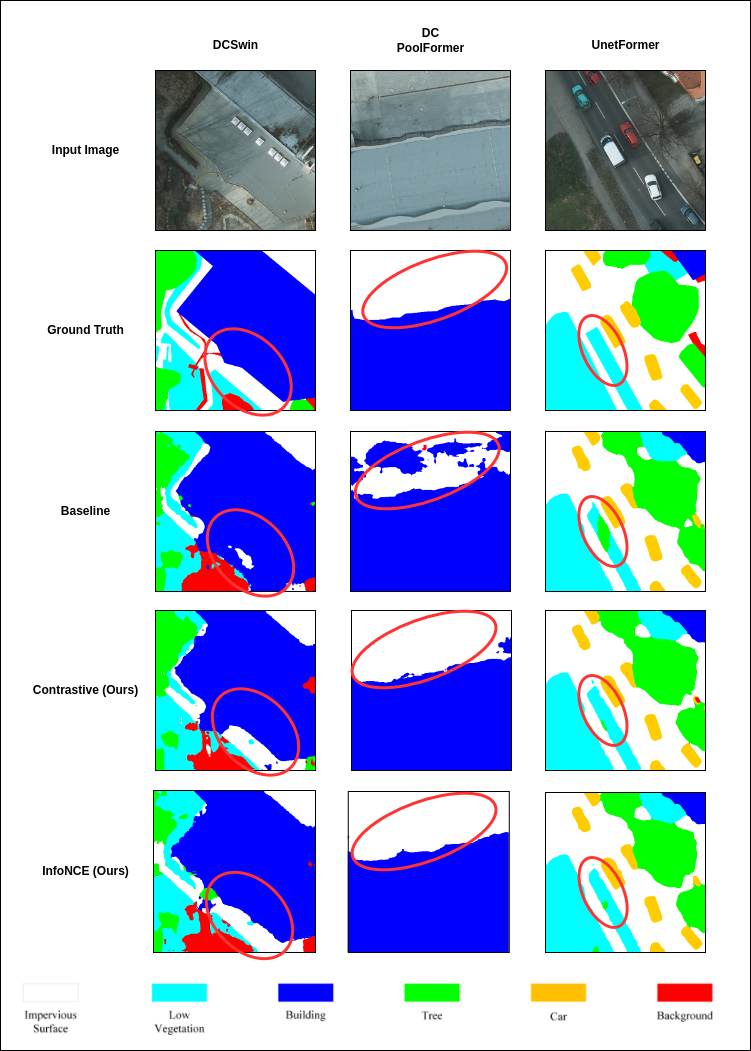}
\caption{Qualitative comparison for all models depicting the enhanced representations from the CT learning scheme. The areas of interest is highlighted with a red circle, showing examples where the difference between baseline and CT is prominent. All visual samples for the comparison has been gathered from the best run across three distinct runs.}
    \label{fig:visualized_results}
\end{figure}

We compare our findings with baseline results for all models, distinguished only by adding the CT learning scheme. We evaluate CT utilizing InfoNCE and CL, both widely used loss functions for image-based contrastive learning. Table \ref{tab:results} enumerates the experimental results, consistently demonstrating advancements over the baselines with at least one of the contrastive loss functions for each model. Fig. \ref{fig:visualized_results} provides a qualitative comparison across all models, illustrating the enhanced semantic representations achieved via CT.

\begin{table*}
\scriptsize
  \begin{center}
    \caption{These are the results from evaluating the ISPRS Potsdam dataset for all models. Values in the 'Mean IoU' column display the average IoU across all runs and classes with the standard deviation in the row below. The values in the class columns are the average IoU for each class across all runs. The results show that CT achieves comparable or better mean IoU for all model types. The best mean IoU for each model is bolded, while the highest IoU for each class is underlined.}
    \label{tab:results}
    \begin{tabular}{c|ccc|ccc|ccc}
      \toprule
      
      \emph{Model} & & DS & & & UF & & & PF & \\
      \hline
      \emph{Backbone} & & Swin & & & Swin & & & PoolFormer & \\
      \hline
      \emph{Source} & \cite{wang2022aimages} & \textbf{Ours} & Ours & \cite{wang2022aimages} & Ours & \textbf{Ours} & \cite{wang2022aimages,Yu2022MetaFormerVision} & Ours & \textbf{Ours} \\
      \hline
      \emph{Contrastive} & - & \textbf{CL} & InfoNCE & - & CL & \textbf{InfoNCE} & - & CL & \textbf{InfoNCE}\\
      \hline
      \emph{Mean IoU} & 0.773 & \textbf{0.774} & 0.773 & 0.769 & 0.768 & \textbf{0.772} & 0.756 & 0.762 & \textbf{0.763}\\
      \hline
      \emph{Standard deviation} & 0.002 & 0.002 & 0.001 & 0.001 & 0.002 & 0.001 & 0.002 & 0.006 & 0.006\\
      \hline
      \emph{Surface} & 0.828 & \underline{0.830} & 0.826 & 0.829 & 0.828 & \underline{0.830} & 0.811 & \underline{0.816} & 0.813 \\
      \hline
      \emph{Building} & 0.885 & \underline{0.890} & 0.886 & 0.889 & 0.887 & \underline{0.890} & 0.869 & 0.873 & \underline{0.874} \\
      \hline
      \emph{Vegetation} & \underline{0.731} & 0.728 & 0.726 & 0.722 & 0.720 & \underline{0.727} & \underline{0.708} & 0.707 & 0.708 \\
      \hline
      \emph{Tree} & 0.729 & 0.727 & \underline{0.731} & 0.719 & 0.715 & \underline{0.722} & 0.705 & \underline{0.711} & 0.708 \\
      \hline
      \emph{Car} & 0.691 & \underline{0.695} & 0.692 & 0.684 & 0.687 & \underline{0.690} & 0.685 & \underline{0.689} & 0.686 \\
      \bottomrule
    \end{tabular}
  \end{center}
\end{table*}

The results show increased Contrastive Transformer (CT) performance in the car class. Intriguingly, this achievement comes despite the car class often being too small to fill the smallest training patches. Nevertheless, CT outperforms baseline models, suggesting that its capacity to create robust semantic representations does not strictly depend on homogenous class representation within individual patches. The results indicate that the CT model successfully generates superior representations compared to the baseline only using negative samples of that class. This showcases the model's adaptability and capacity to form good class representations.

It is important to highlight the flexibility of CT, as it pairs effectively with two different transformer backbones, notably the Swin Transformer and PoolFormer. Our findings show a larger improvement for CT when paired with the PoolFormer backbone. It indicates that the SwinTransformer is more capable of representation learning than PoolFormer. However, this observation warrants further investigation. In light of these findings, we summarize as follows:

\begin{itemize}
    \item CT demonstrates compatibility with diverse backbones and outperforms the baseline models in nearly all conducted experiments.
    \item The model shows an interesting resilience in cases where instances of cars are too small to be considered as positive samples during contrastive learning, suggesting that their inclusion in negative samples is sufficient for CT to enhance the class representation.
    \item Future work should delve deeper into the exploration of which transformer-based backbones are optimally suited to leverage CT's capabilities.
\end{itemize}

\section{Conclusion}
\label{sec:conclusion}
In this work, we introduced the Contrastive Transformer (CT), a novel patch-based contrastive learning scheme for transformers, demonstrating its potential in aerial image segmentation. Our experiments show consistent improvements in segmentation results across all models on the ISPRS Potsdam dataset, with an increase in IoU by 0.7\%. The simplicity and adaptability of the CT model can easily be added to transformer-based models used for urban planning, disaster management, and other fields that rely on accurate aerial image segmentation. Additionally, the technique applies to future transformer-based architectures and can increase their performance. There is substantial potential for extending the CT approach to other image segmentation tasks. Additionally, exploring the impacts of different sampling strategies or refining the contrastive learning process could provide additional performance improvements. The CT model lays a robust foundation for future innovations in this field.

\section{Future Work}
\label{sec:future_work}
The Contrastive Transformer is a novel mechanism that leverages the inherent patch characteristics of vision-based transformers. The results demonstrate the potential to improve the performance and robustness of transformers for dense prediction tasks. Using innate Transformer patches for contrastive learning is simple yet novel and warrants further exploration. This research highlights several avenues for future investigation:

\begin{itemize}
    \item Investigating the impact of patch size and the number of patches used in contrastive learning on CT's performance for dense prediction tasks.
    \item Exploring various augmentation techniques on patches to improve the quality of learned representations.
    \item Developing new architectures that can better leverage the representations learned by contrastive learning.
    \item Studying the effects of different objective functions on the quality of learned representations.
    \item Investigating using unsupervised pre-training techniques to initialize the model before fine-tuning it for specific dense prediction tasks.
    \item Evaluating the generalization capability of learned representations across different datasets and domains.
    \item Exploring the use of state-of-the-art sampling techniques in CT.
    \item Preliminary studies indicate that using attention for patch selection could be useful, with work from \cite{Xia2022VisionAttention} serving as a key accelerator.
    \item Examining the use of the heterogenous patches could harvest great potential as hard samples are often present around the edges of two semantic classes.
    \item Explore the use of CT for other dense prediction tasks such as object detection and instance segmentation.
\end{itemize}

{\small
\bibliography{references}

\begin{thebibliography}{10}

\bibitem{Alonso2021Semi-SupervisedBank}
I.~Alonso, A.~Sabater, D.~Ferstl, L.~Montesano, and A.~C. Murillo.
\newblock {Semi-Supervised Semantic Segmentation With Pixel-Level Contrastive
  Learning From a Class-Wise Memory Bank}, 2021.

\bibitem{Audebert2018BeyondNetworks}
N.~Audebert, B.~Le~Saux, and S.~Lef{\`{e}}vre.
\newblock {Beyond RGB: Very high resolution urban remote sensing with
  multimodal deep networks}.
\newblock {\em ISPRS Journal of Photogrammetry and Remote Sensing}, 140:20--32,
  6 2018.

\bibitem{Chen2020ARepresentations}
T.~Chen, S.~Kornblith, M.~Norouzi, and G.~E. Hinton.
\newblock {A Simple Framework for Contrastive Learning of Visual
  Representations}.
\newblock {\em ArXiv}, 2020.

\bibitem{Dao2021Multi-LabelLearning}
S.~D. Dao, E.~Zhao, D.~Phung, and J.~Cai.
\newblock {Multi-Label Image Classification with Contrastive Learning}.
\newblock {\em ArXiv}, 2021.

\bibitem{Dosovitskiy2021AnScale}
A.~Dosovitskiy, L.~Beyer, A.~Kolesnikov, D.~Weissenborn, X.~Zhai,
  T.~Unterthiner, M.~Dehghani, M.~Minderer, G.~Heigold, S.~Gelly, J.~Uszkoreit,
  and N.~Houlsby.
\newblock {An Image is Worth 16x16 Words: Transformers for Image Recognition at
  Scale}.
\newblock In {\em Proceedings of the 9th International Conference on Learning
  Representations (ICLR)}, pages 1--21, 10 2021.

\bibitem{DumanKeles2023OnSelf-Attention}
F.~Duman~Keles, P.~Mahesakya~Wijewardena, C.~Hegde, S.~Agrawal, and F.~Orabona.
\newblock {On The Computational Complexity of Self-Attention}, 2 2023.

\bibitem{He2020MomentumLearning}
K.~He, H.~Fan, Y.~Wu, S.~Xie, and R.~Girshick.
\newblock {Momentum Contrast for Unsupervised Visual Representation Learning},
  2020.

\bibitem{Huang2022ARECOGNITION}
L.~Huang, S.~Cai, Y.~Zhuang, C.~Jing, Y.~Huang, X.~Tu, and X.~Ding.
\newblock {A TWO-STAGE CONTRASTIVE LEARNING FRAMEWORK FOR IMBALANCED AERIAL
  SCENE RECOGNITION}.
\newblock {\em ICASSP, IEEE International Conference on Acoustics, Speech and
  Signal Processing - Proceedings}, 2022-May:3518--3522, 2022.

\bibitem{isprs.orgISPRSDataset}
{isprs.org}.
\newblock {ISPRS Potsdam Dataset}.

\bibitem{Khoshboresh-Masouleh2020MultiscaleSensors}
M.~Khoshboresh-Masouleh, F.~Alidoost, and H.~Arefi.
\newblock {Multiscale building segmentation based on deep learning for remote
  sensing RGB images from different sensors}.
\newblock {\em Journal of Applied Remote Sensing}, 14(03):1, 7 2020.

\bibitem{Li2022UniFormer:Recognition}
K.~Li, Y.~Wang, J.~Zhang, P.~Gao, G.~Song, Y.~Liu, H.~Li, and Y.~Qiao.
\newblock {UniFormer: Unifying Convolution and Self-attention for Visual
  Recognition}.
\newblock 1 2022.

\bibitem{Li2020InstanceKeypoints}
Q.~Li, L.~Mou, Y.~Hua, Y.~Sun, P.~Jin, Y.~Shi, and X.~X. Zhu.
\newblock {Instance Segmentation of Buildings Using Keypoints}.
\newblock {\em IGARSS 2020 - 2020 IEEE International Geoscience and Remote
  Sensing Symposium}, pages 1452--1455, 9 2020.

\bibitem{Li2021JointSegmentation}
W.~Li, W.~Zhao, H.~Zhong, C.~He, and D.~Lin.
\newblock {Joint Semantic-geometric Learning for Polygonal Building
  Segmentation}.
\newblock {\em 35th AAAI Conference on Artificial Intelligence, AAAI 2021},
  3A:1958--1965, 2021.

\bibitem{Liu2020DenseClassification}
Q.~Liu, M.~Kampffmeyer, R.~Jenssen, and A.~B. Salberg.
\newblock {Dense dilated convolutions merging network for land cover
  classification}.
\newblock {\em IEEE Transactions on Geoscience and Remote Sensing},
  58(9):6309--6320, 9 2020.

\bibitem{Liu2021BootstrappingContrast}
S.~Liu, S.~Zhi, E.~Johns, and A.~J. Davison.
\newblock {Bootstrapping Semantic Segmentation with Regional Contrast}.
\newblock 4 2021.

\bibitem{Liu2018SemanticNetwork}
Y.~Liu, B.~Fan, L.~Wang, J.~Bai, S.~Xiang, and C.~Pan.
\newblock {Semantic labeling in very high resolution images via a self-cascaded
  convolutional neural network}.
\newblock {\em ISPRS Journal of Photogrammetry and Remote Sensing}, 145:78--95,
  11 2018.

\bibitem{Liu2022SwinResolution}
Z.~Liu, H.~Hu, Y.~Lin, Z.~Yao, Z.~Xie, Y.~Wei, J.~Ning, Y.~Cao, Z.~Zhang,
  L.~Dong, F.~Wei, and B.~Guo.
\newblock {Swin Transformer V2: Scaling Up Capacity and Resolution}, 2022.

\bibitem{Matei2008BuildingData}
B.~C. Matei, H.~S. Sawhney, S.~Samarasekera, J.~Kim, and R.~Kumar.
\newblock {Building segmentation for densely built urban regions using aerial
  LIDAR data}.
\newblock {\em 2008 IEEE Conference on Computer Vision and Pattern
  Recognition}, 2008.

\bibitem{Ronneberger2015U-net:Segmentation}
O.~Ronneberger, P.~Fischer, and T.~Brox.
\newblock {U-net: Convolutional networks for biomedical image segmentation}.
\newblock {\em Lecture Notes in Computer Science (including subseries Lecture
  Notes in Artificial Intelligence and Lecture Notes in Bioinformatics)},
  9351:234--241, 2015.

\bibitem{Shi2019BuildingEmbedding}
Y.~Shi, Q.~Li, and X.~X. Zhu.
\newblock {Building segmentation through a gated graph convolutional neural
  network with deep structured feature embedding}.
\newblock {\em Isprs Journal of Photogrammetry and Remote Sensing},
  159:184--197, 1 2019.

\bibitem{vandenOordDeepMind2018RepresentationCoding}
A.~van~den Oord~DeepMind, Y.~Li~DeepMind, and O.~Vinyals~DeepMind.
\newblock {Representation Learning with Contrastive Predictive Coding}.
\newblock 7 2018.

\bibitem{wang2022aimages}
L.~Wang, R.~Li, C.~Duan, C.~Zhang, X.~Meng, and S.~Fang.
\newblock {A Novel Transformer Based Semantic Segmentation Scheme for
  Fine-Resolution Remote Sensing Images}.
\newblock {\em IEEE Geoscience and Remote Sensing Letters}, 19, 2022.

\bibitem{Wang2022UNetFormer:Imagery}
L.~Wang, R.~Li, C.~Zhang, S.~Fang, C.~Duan, X.~Meng, and P.~M. Atkinson.
\newblock {UNetFormer: A UNet-like transformer for efficient semantic
  segmentation of remote sensing urban scene imagery}.
\newblock {\em ISPRS Journal of Photogrammetry and Remote Sensing},
  190:196--214, 8 2022.

\bibitem{Wang2021ExploringSegmentation}
W.~Wang, T.~Zhou, F.~Yu, J.~Dai, E.~Konukoglu, and L.~Van~Gool.
\newblock {Exploring Cross-Image Pixel Contrast for Semantic Segmentation},
  2021.

\bibitem{Xia2022VisionAttention}
Z.~Xia, X.~Pan, S.~Song, L.~E. Li, and G.~Huang.
\newblock {Vision Transformer With Deformable Attention}, 2022.

\bibitem{Yu2022MetaFormerVision}
W.~Yu, M.~Luo, P.~Zhou, C.~Si, Y.~Zhou, X.~Wang, J.~Feng, and S.~Yan.
\newblock {MetaFormer Is Actually What You Need for Vision}, 2022.

\bibitem{Zhang2021LookingLearning}
F.~Zhang, P.~Torr, R.~Ranftl, and S.~R. Richter.
\newblock {Looking Beyond Single Images for Contrastive Semantic Segmentation
  Learning}.
\newblock {\em Advances in Neural Information Processing Systems},
  34:3285--3297, 12 2021.

\bibitem{Zhao2021ContrastiveSegmentation}
X.~Zhao, R.~Vemulapalli, P.~A. Mansfield, B.~Gong, B.~Green, L.~Shapira, and
  Y.~Wu.
\newblock {Contrastive Learning for Label Efficient Semantic Segmentation},
  2021.

\end{thebibliography}
}

\end{document}